\documentclass[conference]{IEEEtran}

\usepackage{cite}
\usepackage{amsmath}
\usepackage{algorithmic}
\usepackage{array}
\usepackage{url}
\usepackage{color}
\usepackage{soul}
\usepackage{enumerate}
\usepackage{amssymb}

\ifCLASSINFOpdf
  \usepackage[pdftex]{graphicx}
  \graphicspath{{../pdf/}{../jpeg/}}
\else
  \usepackage[dvips]{graphicx}
  \graphicspath{{../eps/}}
  \DeclareGraphicsExtensions{.eps}
\fi

\ifCLASSOPTIONcompsoc
  \usepackage[caption=false,font=normalsize,labelfont=sf,textfont=sf]{subfig}
\else
  \usepackage[caption=false,font=footnotesize]{subfig}
\fi

\ifCLASSOPTIONcaptionsoff
  \usepackage[nomarkers]{endfloat}
 \let\MYoriglatexcaption\caption
 \renewcommand{\caption}[2][\relax]{\MYoriglatexcaption[#2]{#2}}
\fi

\begin{document}

\title{DEEP LEARNING HYPERSPECTRAL IMAGE CLASSIFICATION USING MULTIPLE CLASS--BASED DENOISING AUTOENCODERS, MIXED PIXEL TRAINING AUGMENTATION, AND MORPHOLOGICAL OPERATIONS}

% author names and IEEE memberships
% note positions of commas and nonbreaking spaces ( ~ ) LaTeX will not break a structure at a ~ so this keeps an author's name from being broken across two lines.
% use \thanks{} to gain access to the first footnote area
% a separate \thanks must be used for each paragraph as LaTeX2e's \thanks
% was not built to handle multiple paragraphs
%

\author{\IEEEauthorblockN{John E. Ball$^{1}$, and
Pan Wei$^{1}$}
\IEEEauthorblockA{$^{1}$Department of Electrical and Computer Engineering, Mississippi State University
}}

%\author{\IEEEauthorblockN{John E. Ball}
%\IEEEauthorblockA{Electrical and Computer Engineering\\
%Mississippi State University \\
%Mississippi State, MS, 39762\\
%Email: jeball@ece.msstate.edu}
%\and
%\IEEEauthorblockN{Pan Wei}
%\IEEEauthorblockA{Electrical and Computer Engineering\\
%Mississippi State University \\
%Mississippi State, MS, 39762\\
%Email: pw541@msstate.edu}}

\maketitle

%________________________________________________________________________
%
% Abstract
%________________________________________________________________________
%
\begin{abstract}

Herein, we present a system for hyperspectral image segmentation that utilizes multiple class--based denoising autoencoders which are efficiently trained. Moreover, we present a novel hyperspectral data augmentation method for labelled HSI data using linear mixtures of pixels from each class, which helps the system with edge pixels which are almost always mixed pixels. Finally, we utilize a deep neural network and morphological hole-filling to provide robust image classification. Results run on the Salinas dataset verify the high performance of the proposed algorithm.

\end{abstract}

% Note that keywords are not normally used for peerreview papers.

\IEEEpeerreviewmaketitle

\section{Introduction}
\label{sec:Introduction}

\noindent Deep learning has achieved remarkable results in the areas of computer vision and object recognition. More recently, in the field of remote sensing, hyperspectral image (HSI) classification has had significant performance improvements using deep learning \cite{Ball2017Comprehensive}. There are, however, many issues that plague HSI deep learning networks. Among these are the lack of training data, noise, and very high data dimensionality \cite{Ball2017Comprehensive}.

To address the problem of noisy data, a denoising autoencoder (DAE) can be utilized. The DAE will not only denoise the signals, but also can be utilized to extract features. We propose using class--based DAEs that aid in discriminating class information. To address the problem of small amounts of training data, a simple, yet effective, augmentation method is proposed. For each class in the training set, random pairs of pixels are chosen and linearly mixed together, where the mixing ratios always have the true class with the majority abundance values. The rationale for this is that all of these pixels represent the chosen class, but there is variability between the pixels. By linearly mixing training data, a simple, yet effective augmentation strategy allows more data in the training dataset. 

Herein, a HSI classification network composed of a deep class--based DAEs is combined with a deep neural network (NN) classifier. A deep NN learns to extract features from the inputs from each output layer of the DAE that discriminate classes. A softmax classifier provides class labels. Finally, morphological hole-filling cleans up pixel--level misclassifications. The contributions of this paper are as follows:\\

\noindent \textbf{Contribution 1}. A state-of-the-art deep learning network for HSI image classification.\\
   
\noindent \textbf{Contribution 2}. A training data augmentation method for labelled HSI data using linear mixtures of pixels from each class.\\

\noindent \textbf{Contribution 3}. A system utilizing class--based denoising autoencoders for enhanced spectral feature extraction.\\

The contents of this paper are as follows: Sec. \ref{sec:Background} discusses current methods for HSI classification. Sec. \ref{sec:ProposedMethod} provides a detailed account of the proposed method. Sec. \ref{sec:Data} discusses that datasets utilized in this paper. Sec. \ref{sec:Results} discusses results. Finally, Sec. \ref{sec:Conclusions} draws conclusions and lists future work.

\section{Background}\label{sec:Background}
\noindent In HSI classification, many recent works have employed deep learning approaches to achieve superior results over more traditional approaches (which often use hand-crafted features). It is well known that HSI data is noisy, and often denoising can aid classification results. An AE is typically an unsupervised system that learns a mapping from the inputs to a latent (or hidden) space and a mapping from the latent space to the output, such that the output is approximately equal to the input \cite{goodfellow2016deep}. A denoising AE (DAE) tries to reproduce a denoised version of the input. Placing multiple DAEs in series creates a stacked DAE (SDAE). 

DAEs have been employed in HSI processing. Xing et al. \cite{xing2015stacked} utilized SDAEs to pretrain the DL network. Fine tuning was performed using logistic regression. Rectified Linear Units (ReLUs) were utilized for data sparsity. The proposed system achieved better results than a (shallow) Support Vector Machine approach. The system was tested on the Indian Pines, Botswana and Pavia University datasets. Tsagkatakis et al. \cite{Tsagkatakis2016deep} utilized stacked sparse AE (SSAEs) to classify multiple labels per pixel (for the case where field sampling is done at higher resolution that airborne data). A Kullback-Liebler divergence--based term is used to promote sparsity. The hyperspectral data analyzed was 242-band Hyperion data. Liu et al. \cite{liu2015hyperspectral} also uses MDAEs for classification. Spectral and spatial constraints are combined to provide a robust classification. Salinas, Salinas-A, Indian Pines, Pavia Centre and Pavia University datasets are analyzed. Pan et al. \cite{Pan2017_RCVANet} develop the so--called ``rolling-guidance filter'' and vertex component analysis, which achieves higher accuracies in spite of small training data. The datasets analyzed are Indian Pines, Pavia University, and Kennedy Space Center.

\section{Proposed Method}
\label{sec:ProposedMethod}

\subsection{System Overview}

\begin{figure*}[ht]
\centering
\includegraphics[width=7in]{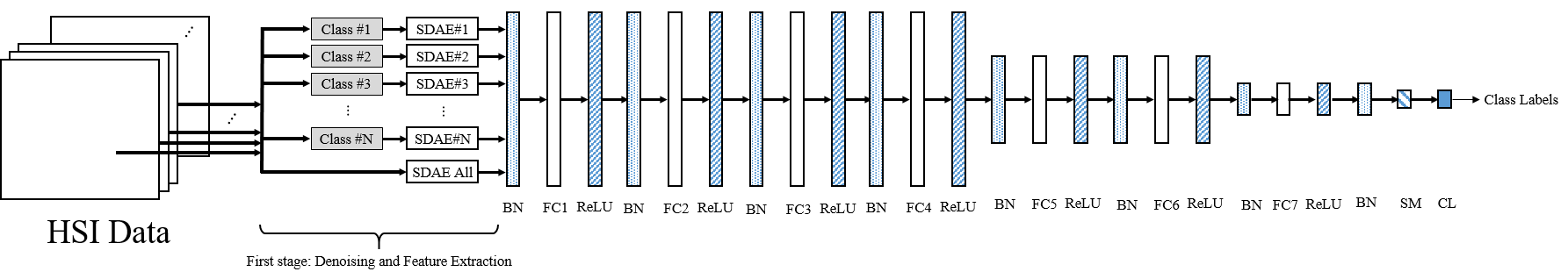}
\begin{center}
(a) \\
\end{center}
\includegraphics[width=7in]{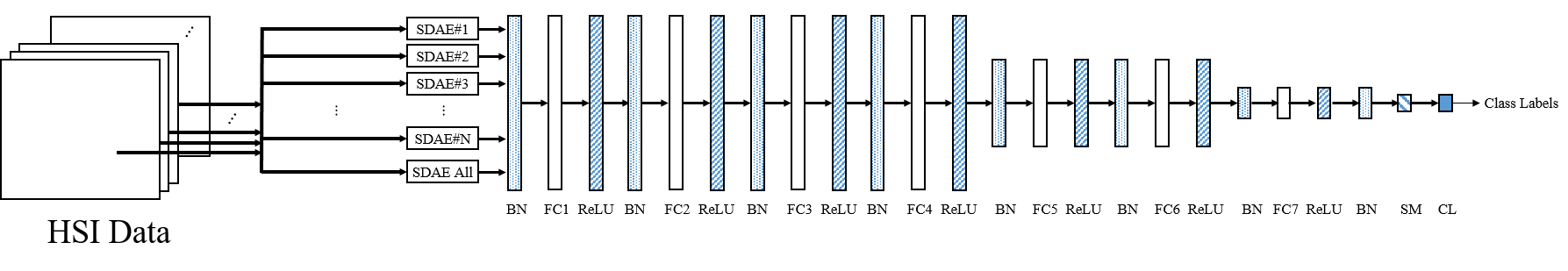}
\begin{center}
(b) \\
\end{center}
\includegraphics[width=2in]{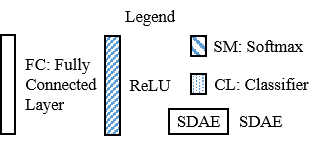}
\begin{center}
(c) \\
\end{center}
\caption{Block diagram of proposed system: (\textbf{a}) Full system in training mode. (\textbf{b}) Full system in testing mode. (\textbf{c}). Legend.}
\label{fig:BlockDiagram}
\end{figure*}

\noindent A HSI classifier utilizing a hybrid deep network consisting of parallel MDAEs and a deep learning NN classifier is proposed. The proposed 24--layer architecture is shown in Figure \ref{fig:BlockDiagram}. The classifier network has two stages: The first stage performs denoising and feature extraction. The second stage performs classification. In the first stage, there are $N$ MDAEs which are class specific. That is, class \#$1$ is trained using training data from class 1. This is repeated for classes $2,3,...,N$ using MDAE\#$2$, MDAE\#$3$, ..., MDAE\#$N$, respectively.

The proposed method utilizes (1) one closed--form solution marginalized denoising autoencoder (MDAE) per endmember, (2) a training data augmentation method which utilizes synthetic linearly mixed pixels to augment the training data, (3) a deep NN to provide an initial classification, and (4) simple morphological processing to provide the final classification. The MDAE outputs, the mean square error (MSE) between the input HSI data and the MDAE outputs, and the original hyperspectral data are all concatenated to form the feature vector for input to the deep NN. Each of these portions of the system are described below.

\subsection{MDAE}

\noindent The MDAE developed by Chen et al. \cite{Chen2012Marginalized} was utilized in this paper. The MDAE has some nice properties: (1) it has a closed form solution, (2) is efficient and easy to implement, (3) it will learn a mapping that provides good reconstruction, and (4) stacking multiple DAEs followed by nonlinear activation functions allows for a rich set of features to be learned automatically. The MMDAE has two parameters, $p \ll 1$, which is the probability that an internal feature is set to zero, and $N_{MDAE}$, the number of layers of the MDAE (non-negative integer). Herein, we use one MDAE layer ($N_{MDAE} = 1$) in order to have a linear mapping.

The input data vector to the MDAE is composed of the hyperspectral signature of each pixel, ${\bf{X}} = \left[ {{{\bf{x}}_1},{{\bf{x}}_2}, \cdots ,{{\bf{x}}_N}} \right] \in {\mathbb{R}^{B \times N}}$, where $B$ is the data dimensionality (number of bands) and $N$ is the number of pixels. The data is presented to the MDAE in a copied version: ${\bf{\bar X}} = \underbrace {\left[ {{\bf{X}},{\bf{X}}, \cdots ,{\bf{X}}} \right]}_{M\;Times}$, where the data is copied $M$ times. Finally, data is presented to the MDAE multiple times with corruptions: ${\bf{\tilde X}}$ is generated from ${\bf{\bar X}}$ by adding Gaussian noise to a subset of randomly selected bands. The MDAE solves for the weight matrix $\textbf{W}$ by minimizing the loss function ${L}\left( {\bf{W}} \right) = trace\left\{ {{{\left( {{\bf{\bar X}} - {\bf{W\tilde X}}} \right)}^T}\left( {{\bf{\bar X}} - {\bf{W\tilde X}}} \right)} \right\}$. This equation has a closed form solution. For more details on the MDAE, please refer to ref. \cite{Chen2012Marginalized}.

\subsection{Data Augmentation}

\noindent To provide improved performance, the training data is augmented creating mixed pixels from the training data. The rationale for this is that pixels on the borders of image regions are usually mixed pixels, and this method will allow more data to be utilized for training. For each class, 25\% of the signatures are randomly selected, and from all other classes, 25\% of signatures are randomly selected. The signatures for each class are paired with signatures from other classes to create mixed pixels. For the given class, the abundance values are always above 55\% to ensure the mixtures are a majority of the given class. A variety of mixing ratios are utilized to provide more training data.

\subsection{Training}

\noindent The network is trained as follows, using a three-step procedure. First, the class specific MDAEs are trained, with each MDAE trained on only one class. Second, the MDAE All is trained on all of the training data. The weights learned by these MDAEs are then kept constant, and the training data is presented to all of the MDAEs.

For each endmember in the training set, MDAE\#$n$ is trained. The training data is corrupted in 40 bands with random Gaussian noise that is zero mean and variance 0.01. Once MDAE\#$n$ is trained, the weights are kept constant. This procedure is repeated for each endmember.

The mean--square--error (MSE) is calculated per class using the MDAE outputs. Each MDAE is trained with one class, and then all data is passed through it. The MSE between the MDAE output and the input signature is then calculated on a per--pixel basis. If the class matches, the MSE is typically lower. 

\subsection{Algorithm Parameters}

\noindent The algorithm parameters are as follows: the probability of MDAE survival is $p = 0.001$, 40 bands are corrupted during MDAE training, the data is repeated $M = 20$ times for the MDAE, $10\%$ of the data is used for training, $25\%$ of the training data is used to create mixed pixels, and the mixing ratios vary in steps of $0.1$. The MDAE outputs and the MSE are concatenated to the data to form the final inputs to the DL NN. The NN stochastic gradient descent training parameters are as follows: The learning rate is $0.04$, the momentum is $0.92$, the mini--batch size is $256$, and the number of training epochs is $20$.

To correct small pixel--based errors, a morphological post--processing step was utilized. Holes were filled in the image on a class--by-class basis.

%
% Proposed Architecture Figure
%

\section{Data}\label{sec:Data}
\noindent Herein, we utilize two publicly available datasets, Salinas \cite{SalinasDataSet}, which is a commonly-used hyperspectral dataset, and Pavia University \cite{PaviaUniversityDataSet}. The Salinas data was collected by the 224-band AVIRIS sensor over Salinas Valley, California, and has high spatial resolution (3.7-meter pixels). The area covered comprises 512 lines by 217 samples. Pre-processing included discarding 20 water absorption bands, namely 108-112, 154-167, and 224. There are 16 classes in this scene, and the several classes are very similar. The Pavia University dataset contains nine classes. The image is 610 rows by 340 columns by 103 bands, with 32,776 labeled pixels and 1.3 meter spatial resolution collected via the ROSIS sensor. Herein, we utilized 4,273 pixels for training, 4,273 for validation and 34,230 for testing.

\section{Results and Discussion}
\label{sec:Results}
\noindent In order to understand how much the different components of the network contribute to the overall results, the following experiments are performed on the Salinas dataset. The experiments are shown in Table \ref{table:DL_Setup}, and are based on different network configurations (different combinations of inputs).  The Salinas dataset was chosen since it had a large number of endmembers and many are very similar. Ten percent of the labeled pixels are used for training, ten percent for validation, and 80 percent for testing.

%
% Experiment Table
%
\begin{table}[h]
\centering
\caption{DL experimental setup.}
\label{table:DL_Setup}
\begin{tabular}{cl}
\textbf{Network} & \multicolumn{1}{c}{\textbf{Description}}          \\
1 & Baseline (only HSI data as inputs). \\
2 & Network 1 + Mixing augmentation. \\
3 & Network 1 + MDAEs. \\
4 & Network 1 + MSE extraction. \\
5 & Network 3 + 4. \\
6 & Network 2 + 3 + 4. \\
7 & Network 6, $p = 0.005$ \\
\end{tabular}
\end{table}

%
% Results tables
%
\begin{table}[h]
\centering
\caption{Testing Results for Salinas dataset. Exp. refers to table \ref{table:DL_Setup}. OA(\%) = Overall Accuracy in percent for test pixels. Highest accuracies are shown in bold font.}
\label{table:Results}
\begin{tabular}{|c|c|c|}
\hline
\textbf{Exp.} & \textbf{\begin{tabular}[c]{@{}c@{}}Raw  OA(\%)\end{tabular}} & \textbf{\begin{tabular}[c]{@{}c@{}}Morph  OA(\%)\end{tabular}} \\ \hline
1 & 93.33 & 96.96 \\ \hline
2 & \textbf{94.55} & 98.20 \\ \hline
3 & 92.76 & 97.50 \\ \hline
4 & 93.12 & 97.55 \\ \hline
5 & 93.15 & 96.82 \\ \hline
6 & 94.05 & \textbf{98.54} \\ \hline
7 & 93.19 & 97.66 \\ \hline 
\end{tabular}
\end{table}

\begin{table}[h]
\centering
\caption{Testing Results for Pavia Universitydataset. Exp. refers to table \ref{table:DL_Setup}. OA(\%) = Overall Accuracy in percent for test pixels. Highest accuracies are shown in bold font.}
\label{table:Results2}
\begin{tabular}{|c|c|c|}
\hline
\textbf{Exp.} & \textbf{\begin{tabular}[c]{@{}c@{}}Raw  OA(\%)\end{tabular}} & \textbf{\begin{tabular}[c]{@{}c@{}}Morph  OA(\%)\end{tabular}} \\ \hline
1 & 93.65 & 96.34 \\ \hline
2 & 93.35 & 96.67 \\ \hline
3 & 92.76 & 96.02 \\ \hline
4 & 93.23 & 95.88 \\ \hline
5 & 93.10 & 95.87 \\ \hline
6 & \textbf{94.38} & \textbf{96.96} \\ \hline
7 & 94.24 & 96.70 \\ \hline 
\end{tabular}
\end{table}

From table \ref{table:Results}, all configurations performed well. The base network solely relied on the hyperspectral data. Network 2 had the best raw overall accuracy at $94.55\%$, while network 6 has the best overall accuracy after morphological processing at $98.54\%$. The mixing augmentation increased accuracy (going from Network 1 to Network 2) by about $1.2\%$, and Networks 3--7 all showed better final performance over the base network. Network 3 added MDAEs, and improved the base result by about $0.5\%$. The best results occurred when the network has mixing augmentation, MDAE inputs and MSE extraction. From table \ref{table:Results2}, all configurations also performed well. Again, the best network was network 6.

Although direct comparison to other methods is not straightforward, the proposed method compares very favorably to results presented in  \cite{xing2015stacked,Tsagkatakis2016deep,liu2015hyperspectral,Pan2017_RCVANet}.

\section{Conclusions}
\label{sec:Conclusions}
\noindent The current system has high accuracy, but only using spectral constraints causes salt--and--pepper type errors (pixel--based errors). Adding a spatial constraint should improve results significantly. In future work, we would like to replace the morphological operations to clean these errors with a CNN-based architecture, which will pincorporate vicinal pixel contextual information.

% Can use something like this to put references on a page
% by themselves when using endfloat and the captionsoff option.
\ifCLASSOPTIONcaptionsoff
  \newpage
\fi

\bibliographystyle{IEEEtran}
\bibliography{refs.bib}

\end{document}